\pdfoutput=1

\documentclass[11pt]{article}
\usepackage[dvipsnames]{xcolor}

\usepackage[final]{coling}
\usepackage{enumitem}

\usepackage{times}
\usepackage{latexsym}

\usepackage[T1]{fontenc}

\usepackage[utf8]{inputenc}

\usepackage{microtype}

\usepackage{inconsolata}

\usepackage{graphicx}

\usepackage{hyperref}
\usepackage{natbib}
\definecolor{darkblue}{rgb}{0, 0, 0.5}
\definecolor{red}{rgb}{0.7, 0.005, 0.15}

\hypersetup{colorlinks=true,citecolor=darkblue, linkcolor=darkblue, urlcolor=darkblue}
\usepackage[nameinlink]{cleveref}

\title{The Only Way is Ethics: A Guide to Ethical Research with Large Language Models}

\author{\textbf{Eddie L. Ungless\textsuperscript{1}},  
\textbf{Nikolas Vitsakis\textsuperscript{2}}, \textbf{Zeerak Talat\textsuperscript{1}}, \textbf{James Garforth\textsuperscript{1}}, \\ \textbf{Björn Ross\textsuperscript{1}}, \textbf{Arno Onken\textsuperscript{1}}, \textbf{Atoosa Kasirzadeh\textsuperscript{1}}, \textbf{Alexandra Birch\textsuperscript{1}}
\\
\textsuperscript{1}University of Edinburgh
\textsuperscript{2}Heriot-Watt University
\\\texttt{a.birch@ed.ac.uk}
}

\begin{document}
\maketitle
\begin{abstract}
There is a significant body of work looking at the ethical considerations of large language models (LLMs): critiquing tools to measure performance and harms; proposing toolkits to aid in ideation; discussing the risks to workers; considering legislation around privacy and security etc. 
As yet there is no work that integrates these resources into a single practical guide that focuses on LLMs; we attempt this ambitious goal. 
We introduce \textsc{LLM Ethics Whitepaper}, which we provide as an open and living resource for NLP practitioners, and those tasked with evaluating the ethical implications of others' work. 
Our goal is to translate ethics literature into concrete recommendations and provocations for thinking with clear first steps, aimed at computer scientists. 
\textsc{LLM Ethics Whitepaper} distils a thorough literature review into clear \textcolor{ForestGreen}{Do's} and \textcolor{red}{Don'ts}, which we present also in this paper. 
We likewise identify useful toolkits to support ethical work. 
We refer the interested reader to the full \textsc{LLM Ethics Whitepaper}, which provides a succinct discussion of ethical considerations at each stage in a project lifecycle, as well as citations for the hundreds of papers from which we drew our recommendations. 
The present paper can be thought of as a pocket guide to conducting ethical research with LLMs. 
\end{abstract}

\section{Introduction}
As LLMs grow increasingly powerful, their advancements in natural language understanding and generation are impressive \citep{min_recent_2023}. 
However, mitigating the risks they present remains a complex challenge, and categorising these risks is a crucial aspect of ethical research related to LLMs~\cite{weidinger2022taxonomy}. Key concerns include the potential to perpetuate and even amplify existing biases present in training data~\citep{gallegos2024bias}, 
challenges in safeguarding user privacy~\citep{yao2024survey}, hallucination or incorrect responses~\citep{abercrombie-etal-2023-mirages, xu2024hallucination}, malicious use of their 
capabilities \citep{cuthbertson_chatgpt_2023}, and infringement of copyright~\citep{Lucchi_2023}. Given that many of these ethical challenges remain unresolved, it is essential for those involved in developing LLMs and LLM-based applications to consider potential harms, particularly as these models see broader adoption.

Several frameworks have already been developed to address AI ethics and safety. For example The U.S. National Institute of Standards and Technology (NIST) has an AI Risk Management Framework (RMF)\footnote{\url{https://www.nist.gov/itl/ai-risk-management-framework}}, which provides broad guidelines for managing AI-related risks. NIST has also recently released a document outlining specific risks and recommended actions for Generative AI\footnote{\url{https://nvlpubs.nist.gov/nistpubs/ai/NIST.AI.600-1.pdf}}. Whilst widely adopted, the NIST guidelines are voluntary. In contrast, the EU AI Act\footnote{\url{https://digital-strategy.ec.europa.eu/en/policies/regulatory-framework-ai}} represents a legally binding regulatory framework designed to ensure the safe and ethical use of AI within the European Union. It emphasises transparency, human oversight, and the prevention of discriminatory outcomes, with the goal of protecting fundamental rights and promoting trustworthy AI.  

The NIST AI RMF and EU AI Act are broad, focusing on AI deployment and risk management across industries. There are other frameworks which are more research-focused, guiding ethical considerations in academic AI work. For example the Conference on Neural Information Processing Systems (NeurIPS) Ethics Guidelines\footnote{\url{https://neurips.cc/public/EthicsGuidelines}} evaluates AI research for ethical concerns as part of the paper submission process. A similar effort from the Association of Computational Linguistics (ACL) has created an Ethics Checklist which guides authors in addressing ethical implications, including limitations, and correct treatment of human annotators.\footnote{\url{https://aclrollingreview.org/responsibleNLPresearch/}}

Despite there being a number of frameworks for the ethical development of AI, we believe that there is still a need for a practical whitepaper focused on the needs of a practitioner working with LLMs. To meet this need, we have created our Ethics Whitepaper\footnote{\url{https://doi.org/10.48550/arXiv.2410.19812}}  \citep{ungless_ethics_2024}, henceforth \textsc{LLM Ethics Whitepaper}. 

\textsc{LLM Ethics Whitepaper} presents insight and pointers to the most relevant ethical research, as it relates to each of the steps in the project lifecycle. It provides more detail than the guidelines of NeurIPS and ACL, but is more ``digestible'' and directly applicable to research with LLMs than the NIST frameworks or the EU AI act. We hope our \textsc{LLM Ethics Whitepaper}, and this Overview paper, will prove valuable to all practitioners, whether you are looking for succinct best practice recommendations, a directory of relevant literature, or an introduction to some of the controversies in the field. 

\begin{figure}
    \centering
    \includegraphics[scale=0.4]{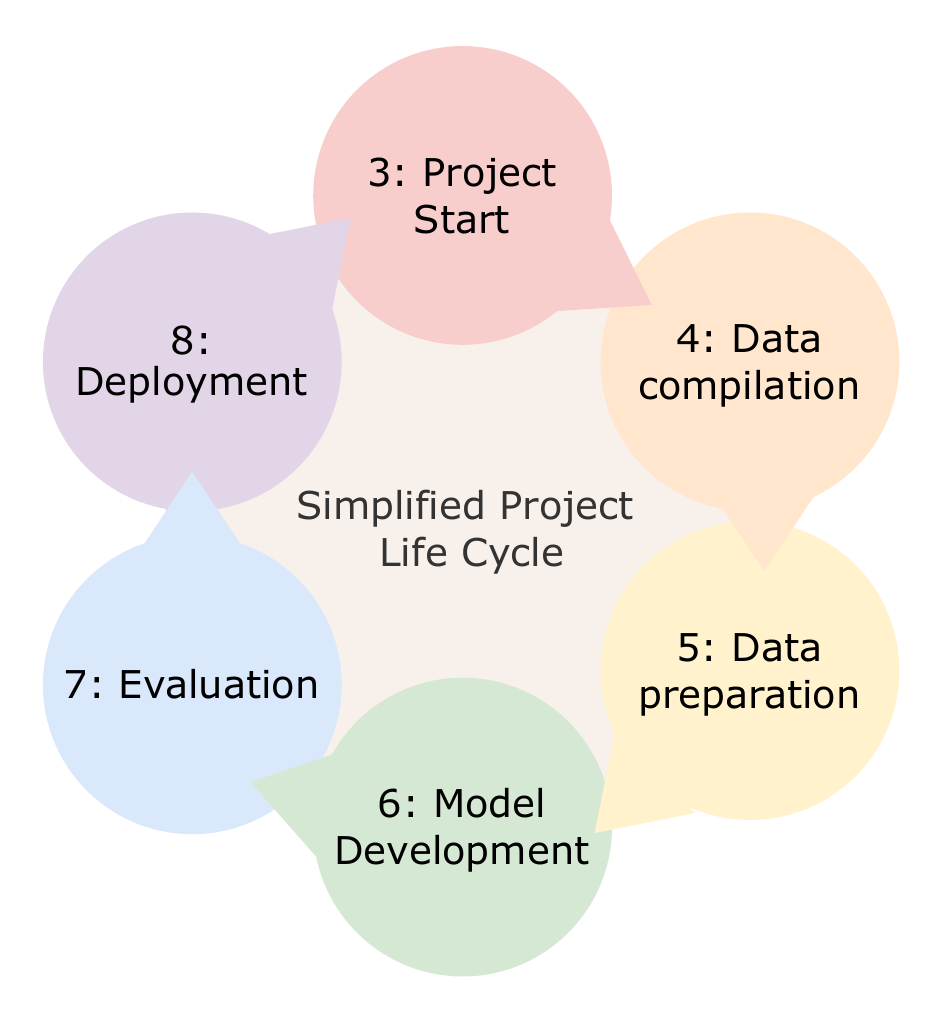}
    \caption{Diagram showing simplified project lifecycle that forms the structure of this paper. This reflects section numbers in the current paper.}
    \label{fig:paperstructure}
\end{figure}

\textsc{LLM Ethics Whitepaper} and this Overview are both structured around a (simplified) project lifecycle, as depicted in Figure \hyperref[fig:paperstructure]{1}. Our aim for these documents is for them to be used as a reference guide throughout a project, rather than for post-hoc reflection. We begin in \Cref{sec:start} by outlining the importance of ethics and discuss themes relevant to the entire development life cycle. An extended version of this Section can also be found in \textsc{LLM Ethics Whitepaper}. In each of the following sections of the present paper we summarise the main topics covered in \textsc{LLM Ethics Whitepaper} and present key resources. Specifically,  \textcolor{ForestGreen}{Do's} and \textcolor{red}{Don'ts} which encompass concrete steps or clear provocations for thinking, which were directly drawn from our extensive literature review, plus tools to guide ethical work. Interested readers should refer to \textsc{LLM Ethics Whitepaper} for more details plus full references, which is hosted on Github\footnote{\url{https://github.com/MxEddie/Ethics-Whitepaper}} to facilitate continual feedback from NLP practitioners. We conclude with Limitations and plans for future developments of \textsc{LLM Ethics Whitepaper}.

\section{Methodology}
A primary goal of \textsc{LLM Ethics Whitepaper} is to provide a comprehensive directory of resources for ethical research related to LLMs. As such, a systematic literature review of the ACL Anthology was conducted (note that \textsc{LLM Ethics Whitepaper} is not itself strictly a systematic literature review paper, see below). The Anthology was searched for paper abstracts containing at least one term from the following key term lists: related to the type of resource = \texttt{tool[a-z]*, toolkit, [A-za-z]*sheets*, guidelines*, principles, framework, approach}; related to ethics = \texttt{ethic[a-z]*, harms*, fair[a-z]*, risks*}. These lists were determined by first using more comprehensive lists then eliminating terms to improve the precision of the search. 
The resulting papers were manually reviewed to determine which were relevant to the scope of \textsc{LLM Ethics Whitepaper}. During the search we identified a 2023 EACL tutorial titled ``Understanding Ethics in NLP Authoring and Reviewing'' \citep{benotti_understanding_2023}. The references for this tutorial were manually reviewed and where relevant included in \textsc{LLM Ethics Whitepaper}. 

A second literature review was conducted using Semantic Scholar using the search terms:  \texttt{toolkit OR sheets OR guideline OR principles OR framework OR approach ethics OR ethical OR harms OR fair OR fairness OR risk AND "language models"}. 
These were likewise manually reviewed for inclusion.

The resulting resources were categorised by their relevance to different stages in a project's lifespan (from ideation to deployment). Primary themes in the literature were identified and used to structure each section; themes were identified by the first author using a bottom-up approach based on the ethical issue(s) each identified paper addressed. Themes were then discussed with all authors and refined in the context of further papers familiar to the authors.

As \textsc{LLM Ethics Whitepaper} progressed, additional resources familiar to the authors were added \textit{ad hoc}. Additionally, papers identified during the research review papers were removed for a number of reasons, either because they were deemed to have limited relevance, or because authors deemed the focus to be too narrow, the recommendations covered by other papers etc. Thus \textsc{LLM Ethics Whitepaper} does not represent our systematic literature review in its entirety, but rather is primarily intended as a practical resource for conducting ethical research related to LLMs, informed by our expertise as practitioners. Combining a literature review with our own expertise ensures broad coverage whilst maintaining a pragmatic focus. 

\section{Project Start}\label{sec:start}

\noindent The social risk of generative AI including LLMs, can have wide-reaching effects from representational harms to safety concerns, which has been widely recognised \cite[see e.g.,][]{weidinger_ethical_2021,bender_dangers_2021,uzun_are_2023,wei_ai_2022}. 
This recognition has given rise to a large number of efforts seeking to evaluate their risks and actualised harms, each effort presenting its own limitations~\cite{Solaiman_Evaluating_2024, goldfarb-tarrant_this_2023, blodgett_language_2020}.
Nevertheless, efforts towards developing technologies that minimise harms, in particular to marginalised communities, are vital.
Best efforts require considering a wide range of topics and questions that must be adapted to each individual application and deployment context. In this Section we explain why ethics is relevant to all practitioners (\Cref{subsec:who}), then highlight resources to aid in the initial discovery process (\Cref{subsec:ideation}). We also lay out best practice that will be valuable to all those working with language technologies, namely related to working with stakeholders, and environmental considerations (\Cref{sec:stakeholders} and \Cref{subsec:energy}).


\subsection{Who needs ethics?}\label{subsec:who}


As computer science becomes pervasive in modern lives, so too does it become intertwined with the experience of those lives. Decisions made by researchers and developers compound together to influence every aspect of the technical systems which ultimately govern how we all live~\cite{winner_artifacts_1980}. This is often at a scale, or level of complexity, which makes it impossible to seek clear resolutions when outcomes are harmful \citep{van2020embedding, kasirzadeh2021reasons, miller2021technology, birhane_values_2022, santurkar2023whose, pistilli2024civics}. 

Techno-cultural artefacts, such as LLMs, have political dimensions \cite{winner_artifacts_1980}, because they further entrench certain kinds of power e.g. marginalised peoples' data is often used without consent or compensation; technology typically works best for language varieties associated with whiteness \citep{blodgett_racial_2017}; benchmarks are published which are biased against minorities \citep{buolamwini_gender_2018}. Unfortunately, the training and work cultures of computer scientists can condition us to believe we can ignore power relations \citep{malazita2019infrastructures}, because the ``objective'' nature of our work seems to absolve us of having to consider issues of our technologies in the world \citep{talat_disembodied_2021} -- when dealing with code and numbers it becomes easier to forget about the real humans who are impacted by our design choices. 
LLMs are no exception \citep{leidner_ethical_2017}, though their recent rise in prevalence has made their ethical dimensions more salient (and more vital to address). 

The design of techno-cultural artefacts like LLMs should be considered interdisciplinary by its very nature, as it requires an understanding of the physical and social systems that they must interact with in order to achieve their function. Experts exist in all of these other areas of study, as well as their intersections, but very often our lack of appreciation for their expertise, or lack of shared language, impede us from seeking them out. This is especially true for expertise in the social sciences and philosophy \citep{raji2021you, inie_idr_2021, danks2022digital}.  

There is a tendency to assume that social and ethical issues are not designers' problems but someone else's \citep{widder2023dislocated}, but this is not the case. If you do not reflect on your design decisions as you make them then you are complicit in the avoidable consequences of those decisions \citep{talat_disembodied_2021}. The decision to follow a code of ethics \cite{mcnamara2018does} or employ a pre-packaged ethical toolkit, does not immediately solve the problem because these decisions require a level of ethical reflection to be effective \citep{wong_seeing_2023}.

\subsection{Laying the Groundwork}\label{subsec:ideation}

It is important to think about ethics from the very beginning, in order to be able to question all aspects of the project, including if specific tasks should even be undertaken.  One way of doing this is by using ethics sheets \citep{mohammad_ethics_2022}, which are sets of questions to ask and answer before starting an AI project. It includes questions like ``Why should we automate this task?'', and ``How can the automated system be abused?''. 

An alternative is using the Assessment List for Trustworthy Artificial Intelligence (ALTAI)\footnote{\url{https://ec.europa.eu/newsroom/dae/document.cfm?doc_id=68342}}, which is a tool that helps business and organisations to self-assess the trustworthiness of their AI systems under development. The European Commission appointed a group of experts to provide advice on its artificial intelligence strategy and they translated these requirements into a detailed Assessment List, taking into account feedback from a six month long piloting process within the European AI community. 
You could use question sets such as these to ensure ethical considerations are present from the start of your project. 

Regulated industries such as aerospace, medicine and finance have critical safety issues to address, and a primary way these have been addressed is using auditable processes throughout a project. Audits are tools for interrogating complex processes, to determine whether they comply with company policy and industry standards or regulations~\citep{liu2012enterprise}. \citet{raji_closing_2020} introduce a framework for algorithmic auditing that supports artificial intelligence system development, which is intended to contribute to closing the gap between principles and practice. 
A formal process such as this can help by raising awareness, assigning responsibility, and improving consistency in both procedures and outcomes \citep{leidner_ethical_2017}. At the very least, your organisation should establish an ethics review board to evaluate new products, services, or research plans.

\subsection{Stakeholders}\label{sec:stakeholders}

\noindent Given the vast amounts of training data required and the wide-reaching applications of LLMs, every project will have many stakeholders e.g. those who provide the data \citep{havens_situated_2020}, end-users of the application \citep{yang2023impact}, or those a model will be used \emph{on}, who are often given limited power to influence design decisions e.g. migrants \citep{nalbandian_eye_2022}. A vital early step is identifying key direct stakeholders and establishing the best ways to work with them in order to build systems that are widely beneficial -- which will be highly context dependent. The ideation toolkit ~\citep{sloane_participation_2022} will help you to identify stakeholders, and can be used alongside existing taxonomies e.g. \citep[i.a.]{lewis_global_2020,langer_what_2021, bird_typology_2023, havens_situated_2020}. 
Crucially, stakeholders should be identified before development, so they can (if they wish) be involved in co-production -- and object to proposed technologies. \citet{kawakami_situate_2024} present a toolkit for early stage deliberation with stakeholders which includes question prompts, while \citet{caselli_guiding_2021} provide 9 guiding principles for effective participatory design -- design which involves mutual learning between designer and participant -- in the context of NLP research. 

You must consider power relations between stakeholders \citep{havens_situated_2020}, and also between yourself and the stakeholders. 
Reflexive considerations about a researcher's own power are rare in computer science research \citep{ovalle_factoring_2023, devinney_theories_2022} but can help establish the limitations of your work \citep{liang_embracing_2021,liang_reflexivity_2021}. 

When working on technologies for indigenous and endangered languages, sensitive stakeholder collaboration is particularly important \citep{bird_decolonising_2020, liu_not_2022, mahelona_openais_2023}. 
Work on stakeholder engagement in NLP can learn much from the Indigenous Data Sovereignty movement \citep{sloane_participation_2022}. 

\subsection{Energy Consumption}\label{subsec:energy}

\noindent Throughout the life cycle of a project, you should consider the energy consumption of your model, which relates to data sourcing practices, model design, choice of hardware, and use at production. \citet{strubell_energy_2019} suggest that model development likely contributes a ``substantial proportion of the... emissions attributed to many NLP researchers''. 
\citet{strubell_energy_2019} call for more research on computationally efficient hardware and algorithms, and the standardised calculation and reporting of finetuning cost-benefit assessments, so researchers can select efficient models (models that are responsive to finetuning) to work with.
Similar recommendations are made by \citet{henderson_towards_2020}, who also provide a framework for tracking energy, compute and carbon impacts. \citet{patterson_carbon_2022} provide best practice for reducing the carbon footprint, including the development of sparse over dense model architectures and the use of cloud computing that relies on renewable energy sources. 
\citet{bannour_evaluating_2021} provide a taxonomy of tools available to measure the impact of NLP technologies. Sasha Luccioni and colleagues have in particular championed the accurate reporting of the carbon emissions of ML systems including LLMs 

\subsection{Key Resources}
Do's and Don'ts
\begin{itemize}[noitemsep]
    \item \textcolor{ForestGreen}{\textbf{Do}} engage with affected communities from the beginning -- \textcolor{red}{\textbf{don't}} just ask for their feedback
    \item \textcolor{ForestGreen}{\textbf{Do}} allow for flexibility in project direction as informed by stakeholder input -- \textcolor{red}{\textbf{don't}} assume what communities want and need
    \item \textcolor{ForestGreen}{\textbf{Do}} consider the power relations between stakeholders -- \textcolor{red}{\textbf{don't}} forget about the relationships with yourself
    \item \textcolor{ForestGreen}{\textbf{Do}} engage with ethics review boards to ensure oversight, or set one up if necessary -- \textcolor{red}{\textbf{don't}} assume because it's computer science that moral and political values are out of scope
    \item \textcolor{ForestGreen}{\textbf{Do}} create an internal audit procedure to ensure ethical processes are developed and followed -- \textcolor{red}{\textbf{don't}} just leave it to a post-hoc review 
    \item \textcolor{ForestGreen}{\textbf{Do}} consider use of compressed models and cloud resources to minimise energy impact -- \textcolor{red}{\textbf{don't}} assume you need energy intensive models for the best performance
\end{itemize}

\noindent Useful Tool(kit)s: 
\begin{itemize}[noitemsep]
    \item Ethics sheets to discover harms and mitigation strategies -- \citet{mohammad_ethics_2022}
    \item The Assessment List for Trustworthy Artificial Intelligence (ALTAI)\footnote{\url{https://ec.europa.eu/newsroom/dae/document.cfm?doc_id=68342}}
    \item Internal audit framework to ensure that ethical processes are implemented and followed -- \citet{raji_closing_2020}
    \item Value Scenarios framework to identify likely impact of technology -- \citet{nathan_value_2007}
    \item Guiding principles for effective participatory design -- \citet{caselli_guiding_2021}
    \item Best practice for reducing carbon footprint during training -- \citet{patterson_carbon_2022}
    \item Taxonomy of tools available to measure environmental impact of NLP technologies -- \citet{bannour_evaluating_2021}
    \item Software package to estimate carbon dioxide required to execute Python codebase -- \url{https://github.com/mlco2/codecarbon} 
\end{itemize}

\section{Data compilation}\label{sec:comp}

 In this Section of \textsc{LLM Ethics Whitepaper} we discuss best practice for compiling original data sets, and critique typical practices such as the position of data as a raw resource rather than something that is transformed by the act of collection. We discuss best practice for addressing issues of consent and safety which includes distinguishing those who produce data and those featured in the data (``data subjects'') and respecting their potentially distinct rights. We also discuss best practice for sharing or using shared resources such as thorough documentation and using API such as \citet{elazar_whats_2024} to explore large data sets. For a full discussion of this section and following sections, please see the \textsc{LLM Ethics Whitepaper}.\\

\noindent\textbf{Key Resources}\\
Do's and Don'ts
\begin{itemize}[noitemsep]
    \item \textcolor{ForestGreen}{\textbf{Do}} reflect on and document the decisions you make when collecting data -- \textcolor{red}{\textbf{don't}} forget that \textit{how} you collect data transforms it 
    \item \textcolor{ForestGreen}{\textbf{Do}} consider if it is ethical to scrape web content, even for content that is publicly available (e.g., by relying on frameworks of ethical data scraping such as \citet{mancosu_what_2020}) -- \textcolor{red}{\textbf{don't}} crawl content that website creators have indicated should not be crawled (e.g. via \texttt{robots.txt} files)
    \item \textcolor{ForestGreen}{\textbf{Do}} consider the subjects of the data -- \textcolor{red}{\textbf{don't}} just think about the rights of data producers
    \item \textcolor{ForestGreen}{\textbf{Do}} respect copyright and privacy from the beginning -- \textcolor{red}{\textbf{don't}} expect the public to do the work of requesting removal (but give them the option!)
    \item \textcolor{ForestGreen}{\textbf{Do}} provide a datasheet for any data set you produce -- \textcolor{red}{\textbf{don't}} forget to document intended use and limitations
\end{itemize}

\noindent Useful Tool(kit)s: 
\begin{itemize}[noitemsep]
    \item Case study structure to identify who is missing from collected data -- \citet{markl_mind_2022}
    \item Best practice from Indigenous data sovereignty movement -- \citet{walter_indigenous_2021}
    \item Checklist for responsible data collection and reuse - \citet{rogers_just_2021} 
    \item API to explore content of popular massive data sets - \citet{elazar_whats_2024} 
    \item Guidelines to create datasheets - \citet{gebru_datasheets_2020}
\end{itemize}

\section{Data preparation}\label{sec:prep}

In this Section of \textsc{LLM Ethics Whitepaper} we discuss how attempts to clean and filter data can cause harm, even where the intention was to prevent harm! For example, toxicity detection systems, be it word lists or ML models, are typically biased in flagging sentences containing marginalised identity terms as toxic \citep{bender_dangers_2021,rottger_hatecheck_2021,calabrese_aaa_2021}. Each cleaning step should be carefully justified. We also discuss best practice for working with crowd workers, who we recommend be treated as human participants (e.g. research is subject to approval from ethical review board where possible).   
We offer guidance for those designing target label taxonomies, such as carefully consider what is assumed and what is lost through your choice of proxy \citep{guerdan_groundless_2023}. We discuss how to handle disagreement between annotators, which are common in subjective tasks and can reflect ideological differences. \\

\noindent\textbf{Key Resources}\\
Do's and Don'ts
\begin{itemize}[noitemsep]
    \item \textcolor{ForestGreen}{\textbf{Do}} carefully reflect on \textit{whose} data you are excluding when cleaning -- \textcolor{red}{\textbf{don't}} rely on popular tools to give you fair results
    \item \textcolor{ForestGreen}{\textbf{Do}} make explicit what information you are trying to record with your choice of proxy -- \textcolor{red}{\textbf{don't}} forget that labels and proxies are simplifications  
    \item \textcolor{ForestGreen}{\textbf{Do}} work with affected communities to define labels and annotate your data -- \textcolor{red}{\textbf{don't}} forget that harm is subjective, and a spectrum 
    \item  \textcolor{red}{\textbf{Don't}} release low quality data that may be repurposed for evaluation
    \item \textcolor{ForestGreen}{\textbf{Do}} gather information about your annotators -- \textcolor{red}{\textbf{don't}} assume annotators with similar identities will give similar annotations
    \item \textcolor{ForestGreen}{\textbf{Do}} treat crowdworkers as human participants and follow best practice for human participant research, such as collecting informed consent; seek formal ethics (e.g. Institutional review board) approval where applicable -- \textcolor{red}{\textbf{don't}} assume that when using paid annotators you do not need to follow typical ethics procedures
\end{itemize}

\noindent Useful Tool(kit)s: 
\begin{itemize}[noitemsep]
    \item Recommendations for those conducting data filtering -- \citet{hong_whos_2024}
    \item Taxonomy of personal information and best practice for privacy -- \citet{subramani_detecting_2023}
    \item Guidance of selecting proxy labels -- \citet{guerdan_groundless_2023}
    \item Best practice when using identity terms as labels -- \citet{larson_gender_2017}
    \item Detailed overview of risks of using crowdworkers -- \citet{shmueli_beyond_2021}
\end{itemize}

\section{Model Development + Selection}\label{sec:modeldev}

In this Section of \textsc{LLM Ethics Whitepaper} we discuss the ethical ramifications of model design and training, and pre-trained model selection decisions. We echo \citet{hooker_moving_2021} in arguing against the belief that all bias issues stem from data imbalance and explain how subtle model design changes can have big impacts on fairness. We also discuss the limitations of debiasing, as current techniques often make superficial changes \citep{gonen_lipstick_2019}, fail to relate to downstream improvements \citep{steed_upstream_2022}, and can in fact exacerbate harm \citep{xu_detoxifying_2021}. We also briefly touch on alignment techniques, exploring the difficulty of defining human values \citep{kasirzadeh2023conversation,kasirzadeh2024plurality} and of maintaining alignment throughout a project. \\

\noindent\textbf{Key Resources}\\
Do's and Don'ts
\begin{itemize}[noitemsep]
    \item \textcolor{ForestGreen}{\textbf{Do}} consider how subtle changes can improve performance for marginalised people -- \textcolor{red}{\textbf{don't}} assume that all bias comes from data imbalance 
    \item \textcolor{ForestGreen}{\textbf{Do}} use and create model cards for documenting correct and indended uses of models -- \textcolor{red}{\textbf{don't}} assume that a model will be reliable for all populations you might care about
    \item \textcolor{ForestGreen}{\textbf{Do}} test for harm on the deployed model -- \textcolor{red}{\textbf{don't}} test on larger versions before compression as harms can be amplified by this process
    \item \textcolor{ForestGreen}{\textbf{Do}} explore techniques such as finetuning to mitigate harm -- but \textcolor{red}{\textbf{don't}} forget this can introduce new harms, and may not be effective
    \item \textcolor{ForestGreen}{\textbf{Do}} maintain vigilance to ensure alignment is maintained throughout a pipeline -- \textcolor{red}{\textbf{don't}} assume there is only one fixed set of human values 
\end{itemize}

\noindent Useful Tool(kit)s: 
\begin{itemize}[noitemsep]
    \item Very clear explanation of how model design choices impact fairness -- \citet{hooker_moving_2021}
    \item Templates to document ML models including intended use context -- \citet{mitchell_model_2019}
    \item Overview of techniques to mitigate LLM harms -- \citet{kumar_language_2022}
\end{itemize}

\section{Evaluation}\label{sec:eval}

Here we discuss some of the ethical problems that can arise during performance evaluation, due for example to evaluation not being robust. We offer best practice and cautions for effective performance evaluation. For example, we caution that benchmarks are not objective and can encourage chasing scores which do not relate to real world improvements. We also discuss in detail the benefits and limitations of many harm evaluation strategies. Despite the ubiquitous nature of the harms of LLMs \citep{rauh_characteristics_2022, weidinger_ethical_2021}, the study of such harms has yet to be standardised. Attempts often lack rigour \citep{blodgett_language_2020,blodgett_stereotyping_2021,goldfarb-tarrant_this_2023}. We briefly present some popular methods for evaluating harms in LLMs, discuss ethical implications and make recommendations. \\

\noindent\textbf{Key Resources}\\
Do's and Don'ts
\begin{itemize}[noitemsep]
    \item \textcolor{ForestGreen}{\textbf{Do}} pair bias metrics that relate to real world (downstream) harms with human evaluation -- \textcolor{red}{\textbf{don't}} rely on quick, quantitative metrics alone, as evaluation in language generation can be unreliable

    \item \textcolor{ForestGreen}{\textbf{Do}} develop and use benchmarks to evaluate concrete, well-scoped and contextualised tasks -- \textcolor{red}{\textbf{don't}} present benchmarks as markers of progress towards general-purpose capabilities  

    \item \textcolor{ForestGreen}{\textbf{Do}} carefully reflect on what specific harm you are trying to measure and why the methodology you have created or borrowed is appropriate -- \textcolor{red}{\textbf{don't}} assume bias metrics can be re-used in all new contexts

     \item \textcolor{ForestGreen}{\textbf{Do}} use alternatives to benchmarks which attempt to capture broader capabilities and risks e.g. audits (e.g. \citet{buolamwini_gender_2018}), adversarial testing (e.g. \citet{niven-kao-2019-probing}) and red teaming~\citep{ganguli_red_2022}
    
    \item \textcolor{ForestGreen}{\textbf{Do}} involve experts and community members in the evaluation of the models -- \textcolor{red}{\textbf{don't}} rely on your intuitions and initial assumptions alone
    
\end{itemize}

\noindent Useful Tool(kit)s: 
\begin{itemize}[noitemsep]
    \item Tools to facilitate test ideation -- \citet{ribeiro_beyond_2020}
    \item Taxonomy of LLM
evaluations -- \citet{chang_survey_2023} -- in particular Section 3.2 on evaluating robustness, ethics, bias, and trustworthiness
\item Repository of tests for (English) language generation safety -- \citet{dinan_safetykit_2022}
    \item Test suite to identifying exaggerated safety behaviour -- \citet{rottger_xstest_2024}
    \item Taxonomy of tests for safety and trustworthiness in LLMs -- \citet{huang_survey_2023} 
    \item Framework for testing reliability of NLP systems -- \citet{tan_reliability_2021}
    \item Bias tests across hundreds of identities (in English) -- \citet{smith_im_2022}
    \item Framework for addressing Sociotechnical (contextualised) Safety -- \citet{weidinger2023sociotechnical}
\end{itemize}

\section{Deployment}\label{sec:deploy}

In this Section of \textsc{LLM Ethics Whitepaper} we summarise likely harms of LLMs after deployment. We introduce the notion of dual -- both negative and positive -- use of LLMs. We discuss the impact of different deployment strategies and the limitations of guardrails. We explain the ramifications of using LLMs to replace humans. Finally we discuss best practice when disseminating your ideas. 

Herein we provide a summary of our discussion of risks.
In their broad overview of the harms that arise from generative AI, ~\citet{Solaiman_Evaluating_2024} present seven over-arching categories of social harms from technical systems, including representational harms; privacy and data protection; and data and content moderation labour. However, in recognition that these cannot be separated from impacts on society, \citet{Solaiman_Evaluating_2024} also present categories of ``impacts'' on society, such as trustworthiness and autonomy, marginalisation and violence,
the concentration of authority,
and ecosystem and environmental impacts.
\citet{weidinger_ethical_2021} and \citet{kumar_language_2022} have also addressed risks of generative AI. While these sets of authors have focused on generative AI, many of the same concerns---such as bias, stereotypes, and representational harms---have been well documented for other NLP technologies~\cite[e.g.,][]{Anand_Don_2024,Bolukbasi_Man_2016,Davidson_Racial_2019,De-Arteaga_Bias_2019}.\\

\noindent\textbf{Key Resources}\\
Do's and Don'ts
\begin{itemize}[noitemsep]
    \item \textcolor{ForestGreen}{\textbf{Do}} consider integrating watermarking into your generative models --  \textcolor{red}{\textbf{don't}} rely on supervised detection models alone
    \item \textcolor{ForestGreen}{\textbf{Do}} pre-release audits to identify biases and security vulnerabilities \cite{madnani_building_2017} -- \textcolor{red}{\textbf{don't}} put the onus on marginalised people to discover harms
    
    \item \textcolor{ForestGreen}{\textbf{Do}} release LLMs in stages, with an initial release to trusted researchers, followed by a gradual wider release \cite{solaiman_release_2019} -- \textcolor{red}{\textbf{don't}} forget the model will change its own environment in terms of both training data and people's expectations
    \item \textcolor{ForestGreen}{\textbf{Do}} continually monitor post-deployment to assess new risks \cite{anderljung2023frontier} -- \textcolor{red}{\textbf{don't}} count on brittle guardrails to prevent harm
    \item \textcolor{ForestGreen}{\textbf{Do}} consider how AI might enhance human experience of work, as well as performance -- \textcolor{red}{\textbf{don't}} assume LLMs can effectively replace human participants 
    \item \textcolor{ForestGreen}{\textbf{Do}} consider how the public perceive your technology -- \textcolor{red}{\textbf{don't}} contribute to the hype cycle
     
\end{itemize}

\noindent Useful Tool(kit)s: 
\begin{itemize}[noitemsep]
    \item Framework to encourage AI that enhances rather than replaces human performance -- \citet{shneiderman_human-centered_2020}
    \item Overview of harms and ramifications of generative AI technologies -- \citet{Solaiman_Evaluating_2024}
    \item A definition of dual use, and a checklist for consideration in research projects -- \citet{kaffee_thorny_2023}
    \item Documentation methodology for risks of LLMs, that could be adapted to document dual use impact -- \citet{derczynski_assessing_2023}
\end{itemize}

\section{Limitations and Future Directions}
Whilst our \textcolor{ForestGreen}{Do's} and \textcolor{red}{Don'ts} are applicable regardless of model language, some of our recommended resources are specific to English. Further, all language-specific resources we discuss in \textsc{LLM Ethics Whitepaper} are specific to English. This reflects a tendency for evaluation resources to be produced only for English, but also the first authors' lack of familiarity with non-English language resources.  We extend a similar qualifier in our inclusion of ethical resources that reflect a largely Western moral perspective. Similarly, this paper primarily addresses the text modality, and does not cover other modalities like speech and images. 
As \textsc{LLM Ethics Whitepaper} is intended as a living document, we can integrate further non-English, non-Western and non-text based resources in future. 
Moreover, the \textcolor{ForestGreen}{do's} and \textcolor{red}{don'ts} are presented in terms of languages for which large amounts of resources already exist. 
Languages for which few resources exist may need additional consideration in terms of data and data subject safety.

This paper and \textsc{LLM Ethics Whitepaper} do not provide full considerations of the topics covered, but rather serve as syntheses with directions for future reading. 
Moreover, \textsc{LLM Ethics Whitepaper} and this paper are informed by the literature they rely on, and do not claim to cover all topics of relevance for the development of LLM and LLM-based applications. 
The \textcolor{ForestGreen}{Do's} and \textcolor{red}{Don'ts} we have drafted are not intended to be the final rule in LLM design. Further, it is crucial that readers of this and \textsc{LLM Ethics Whitepaper} situate the considerations of harms of their work within the contexts that their tools will be applied in. 
\textsc{LLM Ethics Whitepaper} and the \textcolor{ForestGreen}{Do's} and \textcolor{red}{Don'ts} can be thought of as starting points, which we will revise in response to community feedback and further consideration for the subject matter of each section. We welcome input from practitioners on how to make this resource most useful. We are hosting \textsc{LLM Ethics Whitepaper} on Github to expedite this process. We will periodically update the version available on Arxiv to facilitate scholarship.

\section{Conclusion}
In this paper, we have briefly summarised the topics covered in \textsc{LLM Ethics Whitepaper}, and highlighted topics of particular interest. 
We synthesise arguments that LLMs and LLM-based applications can have large impacts on society, and therefore developers of such systems need to attend to the types of harms they risk, and seek to mitigate such risks. 
Here, and in \textsc{LLM Ethics Whitepaper}, 
we seek to address the gap in resources for conducting ethical research with LLMs that falls between professional association guidelines, and AI frameworks with extremely broad scope, and provide researchers and practitioners with a starting point for their inquiry into ethical research with and development of LLM applications. 

\section*{Acknowledgements}
This work was partly funded by the Generative AI Laboratory (GAIL), University of Edinburgh. Alexandra Birch was partly funded by the UK Research and Innovation (UKRI) under the UK government’s Horizon Europe funding guarantee [grant number 10039436 (Utter)]. Eddie L. Ungless was supported by the UKRI Centre for Doctoral Training in Natural Language Processing, funded by the UKRI (grant EP/S022481/1) and the University of Edinburgh, School of Informatics.

\bibliography{eddiereferences, ethics}

\end{document}